\pdfoutput=1

\documentclass[11pt]{article}

\usepackage[final]{coling}

\usepackage{times}
\usepackage{latexsym}
\usepackage{booktabs}
\usepackage{multirow}
\usepackage{subcaption}
\usepackage{amsmath}

\usepackage[T1]{fontenc}

\usepackage[utf8]{inputenc}

\usepackage{microtype}

\usepackage{inconsolata}

\usepackage{graphicx}

\usepackage{color}

\title{Finetuning LLMs for Comparative Assessment Tasks}

\author{Vatsal Raina, Adian Liusie, Mark Gales \\
        ALTA Institute, University of Cambridge \\ \texttt{\{vr311,al826,mjfg\}@cam.ac.uk}}

\begin{document}
\maketitle

\begin{abstract}
Automated assessment in natural language generation is a challenging task. Instruction-tuned large language models (LLMs) have shown promise in reference-free evaluation, particularly through comparative assessment. However, the quadratic computational complexity of pairwise comparisons limits its scalability. To address this, efficient comparative assessment has been explored by applying comparative strategies on zero-shot LLM probabilities. We propose a framework for finetuning LLMs for comparative assessment to align the model’s output with the target distribution of comparative probabilities. By training on soft probabilities, our approach improves state-of-the-art performance while maintaining high performance with an efficient subset of comparisons.
\end{abstract}

\section{Introduction}

Automatically assessing the quality of texts generated by the natural language generation (NLG) system remains a challenging task \citep{gao2024llm}. A recent approach which has gained considerable popularity is LLM-as-a-judge \cite{zheng2023judging}, where instruction-tuned LLMs are prompted zero-shot to predict the quality of texts generated by other systems. In particular, LLM comparative assessment \cite{liusie2024llm, qin2024large}, where pairs of texts are compared to determine which is better, has demonstrated strong correlations with human judgements, typically better than those from LLM absolute assessment \cite{liu2023g, liusie2024llm}. Naive comparative assessment, though, scales quadratically with the number of items, which can be impractical when deployed to real-world settings. Hence, more recently, efficient comparative assessment \citep{liusie2024efficient} was explored where by using the LLM probabilities within a product-of-experts (PoE) framework, assessment can be achieved using a subset of the possible comparisons.

Beyond the zero-shot domain, recent studies have shown the benefits gained when systems are fine-tuned for bespoke tasks, including for LLM absolute assessment \cite{latif2024fine} and comparative assessment \cite{park2024paireval}. However, the various experts proposed within the PoE fromework (e.g. Bradley-Terry) make strong assumptions about the underlying distribution of the pairwise probabilities. The differences between the true and assumed distributions can limit the benefits of fine-tuning comparative systems using hard decisions. 
Therefore, here, we tackle this distributional mismatch by forcibly training the LLM under the assumed distribution of interest. Specifically, the pairwise difference in training scores are scaled to soft training probabilities under the target distribution. By training the LLM with these soft pairwise probabilities, the true inference time probabilities can be expected to match the assumed distribution in the PoE framework for comparative assessment. We demonstrate the benefits that finetuning in this fashion has for LLM comparative assessment, and our contributions can be summarized as follows: 1. We propose a framework for LLM comparative assessment training; 2. we demonstrate that finetuning with soft comparative probabilities under a target distribution enables higher performance with a highly efficient number of comparisons than finetuning with hard binary decision training.

\section{Related work}

\textbf{LLM Comparative Assessment} Recent research has investigated using LLMs to make pairwise comparisons to rank text outputs, as well as the associated computational efficiency. \citet{qin2024large} use pairwise comparisons to retrieve relevant sources, using both the full comparison set and sorting-based techniques. \citet{liusie2024llm} compute the win-ratio using the sets of possible comparisons, demonstrating that for medium-sized LLMs, pairwise comparisons surpass traditional scoring methods for various NLG assessment benchmarks. They also show that performance declines significantly as the number of comparisons falls. Additionally, \citet{liu2024aligning} emphasize the limitations of LLM scoring, advocating for pairwise comparisons and introducing PAirwise-preference Search (PAIRS), a merge sort variant that leverages LLM probabilities. Finally, \citet{liusie2024efficient} apply a product of experts framework to zero-shot LLM probabilities for higher performing comparative assessment with a subset of comparisons. In this work, we extend existing comparative assessment methods by exploring the finetuning of such systems. 

\textbf{Finetuning Prompted Assessment Systems} \citet{latif2024fine} investigate fine-tuning ChatGPT for absolute assessment \citet{park2024paireval}. \cite{ouyang2022training} use human preferences rankings to train the reward model under the Bradley-Terry model, and \citet{park2024paireval} use the average probability across randomly sampled comparisons as a quality metric and demonstrate performance improvements by supervised training. However, in all these methods, only hard decisions are used to train systems, and they don't consider the impact it has on downstream scoring mechanisms, such as the PoE framework.



\section{LLM comparative assessment}

\subsection{Scoring methods}

For the task of NLG assessment, the objective is to score a set of candidate texts for a selected attribute (e.g. coherency or question complexity). 
Let $x_{1:N}$ denote a set of $N$ candidate texts with corresponding true scores for the attribute of interest, $s_{1:N}$. Let $\mathcal{M}$ be a comparative model that returns the probability of $x_i$ being greater than $x_j$ for the assessed attribute, $p_{ij}$. 

By observing the outcome of a set of pairwise comparisons, $\mathcal{C}_{1:K}$, various methods exist to convert the outcomes to the predicted scores, $\hat{s}_{1:N}$. Following \citet{liusie2024efficient}, we consider several method methods of mapping a set of comparisons to assessment scores. When using hard binary decisions, we use the win-ratio \citep{qin2024large, raina2024question} and the Bradley-Terry model (BT) \citep{bradley1952rank}, while when probabilities are leveraged, we consider equivalent `soft' approaches such as the average probability (avg-prob) \citep{park2024paireval} and the Bradley-Terry experts in the PoE framework\footnote{\citet{liusie2024efficient} also consider Gaussian experts for the PoE framework, but we focus our experiments on the soft Bradley-Terry expert as it performs marginally better than the Gaussian experts.} (PoE-BT).
In PoE-BT, the score difference between a pair of items is assumed to be conditioned on the LLM comparative probability, with the output probability distribution given in Equation \ref{eq:poe-bt}.
\begin{equation}
\small
    \begin{aligned}
        p(s_i - s_j | p_{ij}) =  
        \frac{1}{Z_{ij}} \sigma(s_i - s_j)^{p_{ij}} \left( 1 - \sigma(s_i - s_j) \right)^{1 - p_{ij}}
    \end{aligned}
\label{eq:poe-bt}
\end{equation}
where $Z_{ij}$ is a normalizing constant to ensure a valid pdf and $\sigma(\cdot)$ is the sigmoid function. The predicted scores $\hat{s}_{1:N}$ are then the scores which maximise the PoE probability,
\vspace{-2mm}
\begin{equation}
\hat{s}_{1:N} = \arg\max_{s_{1:N}} \frac{1}{Z} \!\! \prod_{i, j \in \mathcal{C}_{1:K}} \!\!\! 
 {\tt p}(s_i \! - \! s_j | p_{ij}) \\
\end{equation}
\vspace{-4mm}



\begin{table*}[htbp!]
\centering
\small 
\begin{tabular}{lcc|ccc|ccc}
\toprule
\multirow{2}{*}{Comparisons} & \multirow{2}{*}{Model} & \multirow{2}{*}{Mode}  & \multicolumn{3}{c|}{USMLE} & \multicolumn{3}{c}{CMCQRD} \\
& & & $\rho$ $(\uparrow)$ & $r$ $(\uparrow)$ & rmse $(\downarrow)$ & $\rho$ $(\uparrow)$ & $r$ $(\uparrow)$ & rmse $(\downarrow)$ \\
\midrule
\multirow{5}{*}{Full [$O(N^2)$]}
    &  \multirow{2}{*}{GPT4o mini} 
        &  zero-shot & 35.5 & 28.2 & 30.3 & 47.9 & 47.3 & 8.94   \\
    &   &  hard & 73.3 & 68.0 & 23.2 & 53.8 & 54.7 & 8.50 \\
    \cmidrule{3-9}
    &  \multirow{3}{*}{Llama-3.1-8B} 
        &  zero-shot & 26.3 & 28.1 & 30.3 & 12.6 & 13.7 & 10.06    \\
    &   &   hard & 69.3 & 64.4 & 24.2 & 41.9 & 41.2 & 9.25 \\
    &   &   soft & 69.3 & 65.5 & 23.8 & 47.8 & 49.1 & 8.84 \\
\midrule
\multirow{5}{*}{Partial [$4N$]}
    &  \multirow{2}{*}{GPT4o mini} 
        &  zero-shot & 27.8 & 21.5 & 30.9 & 14.5 & 16.7 & 10.00   \\
    &   &  hard & 64.8 & 60.4 & 25.2 & 50.9 & 52.6 & 8.64 \\
    \cmidrule{3-9}
    &  \multirow{3}{*}{Llama-3.1-8B} 
        &  zero-shot & 22.9 & 27.4 & 30.4 & 12.1 & 12.9 & 10.07   \\
    &   &   hard & 59.6 & 56.4 & 26.1 & 41.3 & 39.1 & 9.35 \\
    &   &   soft & 61.3 & 57.4 & 25.9 & 48.1 & 49.3 & 8.83 \\
   \bottomrule
    \end{tabular}
\caption{Results for comparative assessment using PoE-BT as the scoring method.}
\label{tab:results}
\end{table*}

\subsection{Finetuning Systems}
The product of experts perspective assumes a certain distribution to the LLM probabilities. For example, the Bradley-Terry model assumes a sigmoidal distribution. However, zero-shot comparative prompting of LLM systems does not necessarily match the assumed distribution of probabilities. 

If we finetune LLMs for comparative assessment, we have the flexibility to control the distribution of probabilities returned by the comparative model. Hence, we convert a set of training scores to a set of training pairwise probabilities according to:

\begin{equation}
\label{eq:probs}
    p_{ij} = f\left(\frac{s_i-s_j}{\gamma \sigma_s}\right)
\end{equation}
where $\sigma_s$ denotes the standard deviation of the set of training scores; $\gamma$ is hyperparameter controlling the spread of the probabilities (see Appendix \ref{app:impact} for its impact). Note, $\gamma=0$ is equivalent to binary decisions, while large values of $\gamma$ push the probabilities out of the saturation region. In general, we consider $f\in\{ \sigma, \Phi \}$, where $\sigma$ matches the sigmoid distribution for Bradley-Terry, while $\Phi$ is the cumulative distribution function of Gaussians used for Thurstone-Mosteller \citep{handley2001comparative}. We restrict our analysis to just Bradley-Terry (hence $f=\sigma$) as an approximately linear relationship can be established between Bradley-Terry scores and Thurstone-Mosteller scores (see Appendix \ref{app:rel}). Given the set of pairwise probabilities, we train the LLM according to a soft binary cross entropy loss:
\begin{equation}
\label{eq:loss}
    \mathcal{L}(\theta) = - ( y \cdot \log(\hat{y}) + (1-y) \cdot \log(1-\hat{y}) )
\end{equation}
where $y = p_{ij}$ calculated from Equation \ref{eq:probs} as the label while $\hat{y}$ is the prediction from the model.

\section{Experiments}

\subsection{Data}

We consider two datasets: USMLE \citep{yaneva2024findings} and CMCQRD \citep{mullooly2023cambridge, liusie2023analysis}. USMLE is a medical multiple-choice reading comprehension (MCRC) dataset where each item has been annotated with the average response time for candidates answering the question. CMCQRD is an educational MCRC dataset annotated with difficulty scores.

\begin{table}[htbp!]
\centering
\small 
\begin{tabular}{l|ccc}
\toprule
Data & Train & Test & Task \\
\midrule
USMLE & 466 & 201 & response time \\
CMCQRD & 464 & 194 & difficulty \\
   \bottomrule
    \end{tabular}
\caption{Data statistics.}
\label{tab:data}
\end{table}

Table \ref{tab:data} summarizes the main statistics. USMLE consists of 667 items where the standard split has 466 training examples and 201 for testing. All items have unique contexts. CMCQRD has 658 items. With no standard split, we partition the dataset into a training set of 464 training and 194 test examples. There are 78 unique contexts across the whole dataset with no overlap between the train and test splits.
The  USMLE dataset additionally has difficulty scores \footnote{We present our experimental results for this task on USMLE in Appendix \ref{app:usmlediff}.}.  Note, we selected USMLE and CMCQRD for our comparative finetuning experiments as these were the only NLG datasets (to our knowledge) that have human annotated attributes and are sufficiently large to warrant training a comparative system.

\subsection{Models}

\begin{figure*}[htbp!]
    \centering
    \begin{tabular}{cc}
        \begin{subfigure}[b]{0.33\textwidth}
            \centering
            \includegraphics[width=\textwidth]{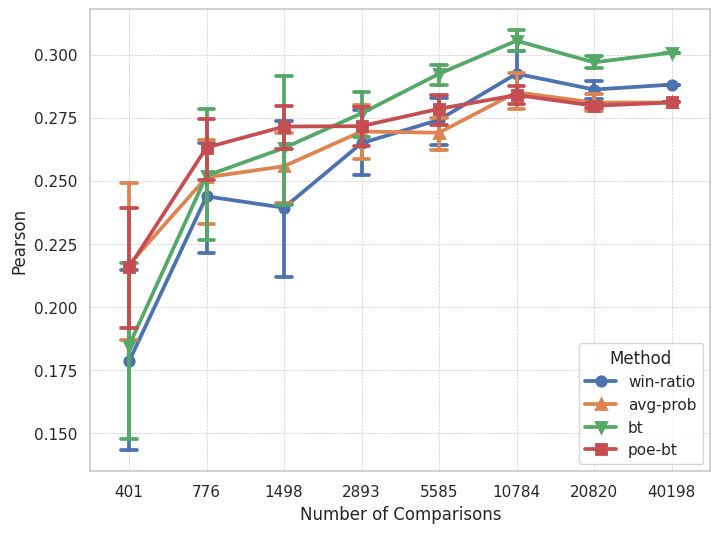}
            \caption{Zero-shot}
            \label{fig:subfig1}
        \end{subfigure} &
        \begin{subfigure}[b]{0.33\textwidth}
            \centering
            \includegraphics[width=\textwidth]{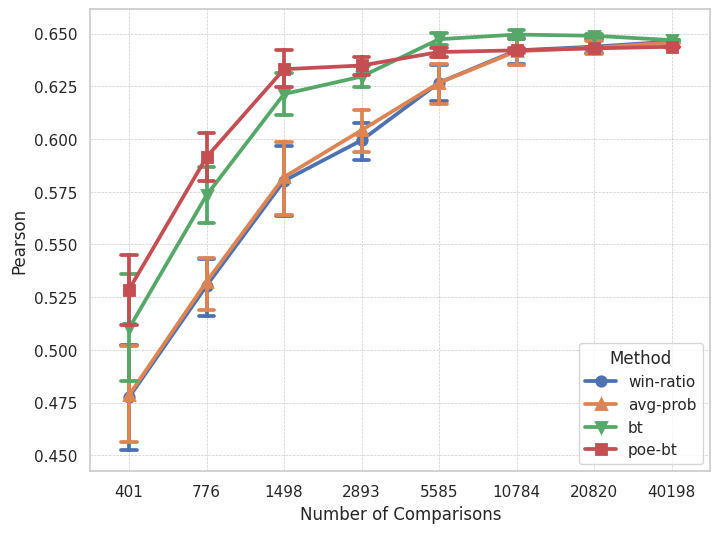}
            \caption{Finetuned hard}
            \label{fig:subfig2}
        \end{subfigure} 
        \begin{subfigure}[b]{0.33\textwidth}
            \centering
            \includegraphics[width=\textwidth]{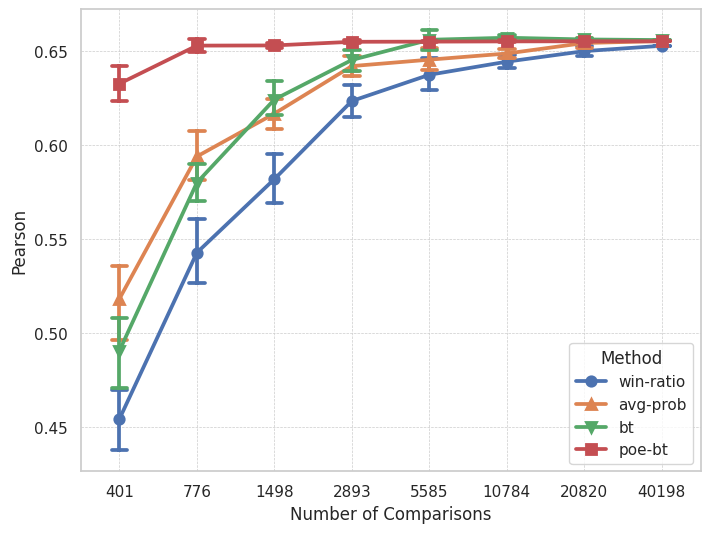}
            \caption{Finetuned soft}
            \label{fig:subfig3}
        \end{subfigure}
    \end{tabular}
    \caption{USMLE response time estimation: Efficient comparisons with Llama-3.1.}
    \label{fig:efficient_resp}
\end{figure*}

\begin{figure*}[htbp!]
    \centering
    \begin{tabular}{cc}
        \begin{subfigure}[b]{0.33\textwidth}
            \centering
            \includegraphics[width=\textwidth]{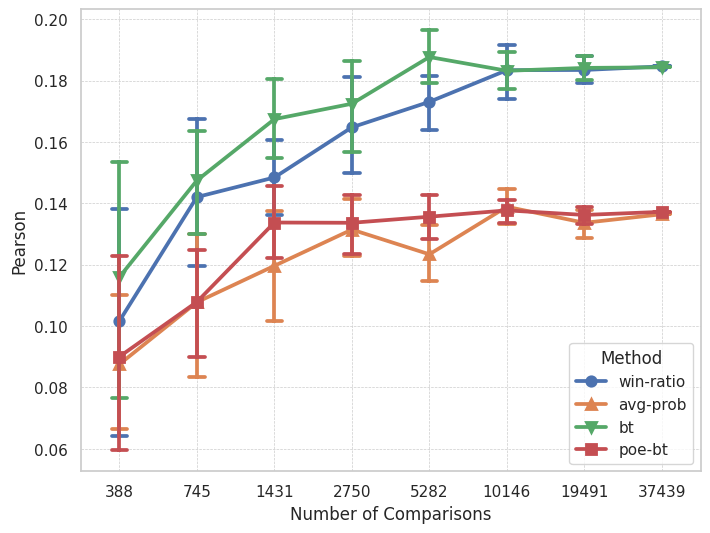}
            \caption{Zero-shot}
            \label{fig:subfig1}
        \end{subfigure} &
        \begin{subfigure}[b]{0.33\textwidth}
            \centering
            \includegraphics[width=\textwidth]{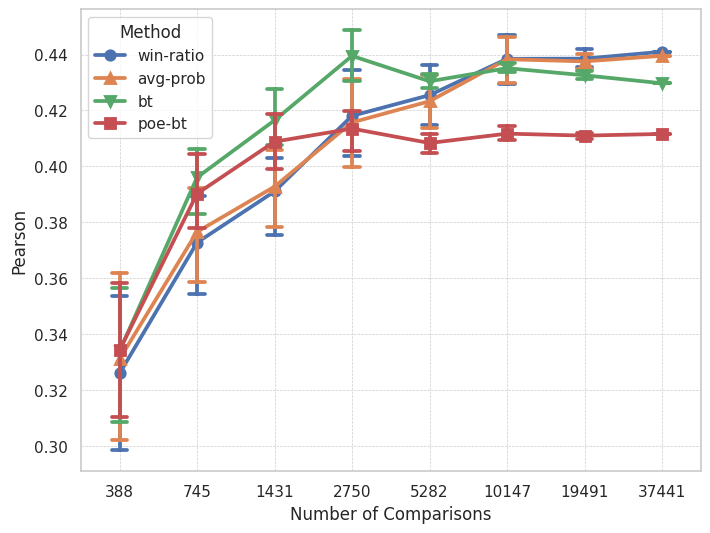}
            \caption{Finetuned hard}
            \label{fig:subfig2}
        \end{subfigure} 
        \begin{subfigure}[b]{0.33\textwidth}
            \centering
            \includegraphics[width=\textwidth]{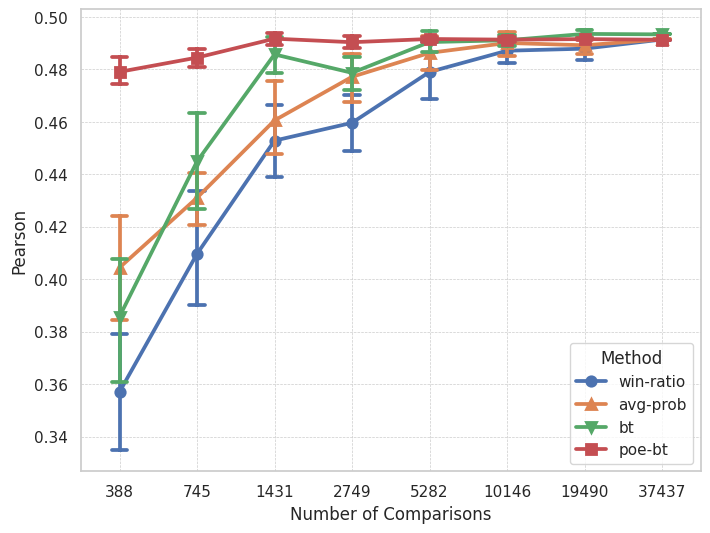}
            \caption{Finetuned soft}
            \label{fig:subfig3}
        \end{subfigure}
    \end{tabular}
    \caption{CMCQRD difficulty estimation: Efficient comparisons with Llama-3.1.}
    \label{fig:efficient_diff}
\end{figure*}

The comparative system $\mathcal{M}$ is an instruction-tuned LLM, configured with an appropriate prompt (e.g. `which item from text 1 or text 2 is better according to the attribute') - see Appendix \ref{app:prompts}. As is common for getting continuous outputs from LLMs \citep{liusie2024efficient}, the LLM logits over the label classes (1 and 2) are taken to calculate $p_{ij}$ for $\mathcal{C}_k$ using softmax.
We run our analysis using GPT4o mini \footnote{Available at: \url{https://openai.com/index/gpt-4o-mini-advancing-cost-efficient-intelligence/}} as a closed-source solution and Llama-3.1-8B \citep{dubey2024llama} as the open-source solution.
For Llama-3.1-8B, we run comparative assessment for zero-shot, hard finetuning ($\gamma=0$) and soft finetuning \footnote{Code and corresponding exact data in splits will be made available.}. The finetuning is based on Equation \ref{eq:loss}, where hard finetuning uses 0 or 1 while soft finetuning uses the soft probabilities from Equation \ref{eq:probs} for the labels. For soft finetuning, we find $\gamma=5.0$ gives us the best results. 
Due to the closed-source access, with GPT4o mini, it is only possible to do hard finetuning. It is interesting that closed-source access training is better designed for comparative than absolute training as the models must be trained to predict an output class (rather than a continuous score). See Appendix \ref{app:hyp} for hyperparameter details.

\section{Results}

Table \ref{tab:results} summarizes the performance of comparative assessment systems. We use Spearman's correlation coefficient, $\rho$, Pearson's correlation coefficient, $r$, and root mean squared error, rmse, between the predicted and true scores on each of the test sets as the performance metrics. Rmse is calculated after linear scaling of the predictions to the true scores. PoE-BT is used for comparative assessment.
As expected, hard finetuning of GPT4o mini substantially boosts performance compared to the zero-shot numbers. A similar improvement from zero-shot performance can be observed when hard finetuning Llama-3.1-8B.

Figures \ref{fig:efficient_resp} and \ref{fig:efficient_diff} present the performance evolution (using Pearson) with an efficient number of comparisons. Hard finetuning leads to improved performance for a small number of comparisons compared to the zero-shot curves. By applying soft finetuning, there is minimal degradation in the PoE-BT curve with an extremely small subset of comparisons. Table \ref{tab:results} further quantifies the benefits of soft finetuning by presenting the results with a partial number of comparisons at an operating point of $4N$, where $N$ is the number of items ($N^2$ is the order of the maximal comparisons.) selecting a high $\gamma$ in soft finetuning pushes the distribution of the pairwise probabilities outside the saturation region of the sigmoid. This means that very few comparisons are needed for each item to deduce the overall ranking as a comparison  between item A and B as well as a comparison between item A and C enables the comparison between item A and C to be inferred.
Table \ref{tab:rt_baselines} further shows that our best comparative system outcompetes all the submitted solutions \citep{rodrigo2024uned, tack2024itec, felice2024british} to the BEA shared task 2024 \citep{yaneva2024findings} for response time.

\begin{table}[htbp!]
\centering
\small 
\begin{tabular}{l|c}
\toprule
Approach & rmse $(\downarrow)$ \\
\midrule
Dummy Regressor Baseline  & 31.7 \\
UNED - run2  & 23.9 \\
ITEC - Lasso  & 24.1 \\
EduTec - roberta  & 25.6 \\
Ours: comparative  & \textbf{23.2} \\
   \bottomrule
    \end{tabular}
\caption{Benchmarking against baselines for USMLE.}
\label{tab:rt_baselines}
\end{table}

\section{Conclusions}

Here, a framework of finetuning LLMs for comparative assessment tasks is proposed. Due to the quadratic compute cost in a full-set of comparisons, it is of high interest to achieve the same assessment performance with an efficient subset of comparisons.
We finetune LLMs in comparative manner using both binary decisions and soft probabilities. The soft probabilities are calculated from the training items' scores using a sigmoid function, enabling the PoE set-up on the Bradley-Terry method of pairwise comparisons to achieve near maximal performance with few comparisons.

\section{Limitations}

We finetune two different LLMs for comparative assessment: GPT4o mini as a closed-source model and Llama-3.1-8B as the open-source model. Both of these models are the smallest in their series of models. Ideally, it would be useful to replicate the experiments using larger models, but there isn't the computational budget available to run larger scale models.

\section{Ethics statement}

There are no ethical concerns with this work.



\bibliography{custom}

\appendix

\newpage
.
\newpage

\section{Impact of $\gamma$}
\label{app:impact}

\begin{figure}[htbp!]
    \centering
            \includegraphics[width=0.48\textwidth]{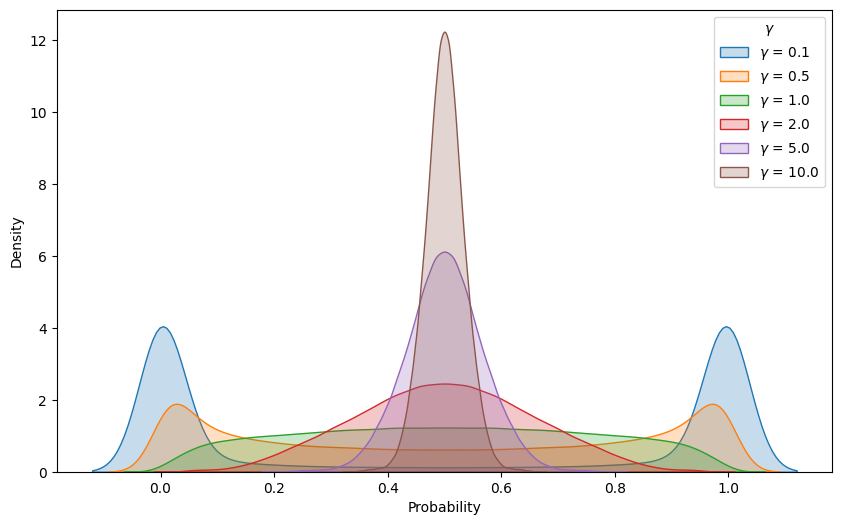}
    \caption{Impact of distribution of training probabilities based on choice of $\gamma$ in sigmoid.}
    \label{fig:gamma_sweep}
\end{figure}

Equation \ref{eq:probs} details the approach used to compute training pairwise probabilities (for the soft cross-entropy loss function) from the true score difference between a pair of items. $\gamma$ in this equation controls the distribution of the probabilities. Let $f(\cdot)=\sigma(\cdot)$. Then, based on the profile of the sigmoid function, a larger $\gamma$ leads to a greater concentration of pairwise probabilities around 0.5. Figure \ref{fig:gamma_sweep} presents the various profiles of the pairwise probabilities computed on the true response time scores of the training split of USMLE. In general, $\gamma=0$ leads to operating in the saturation region of the sigmoid and hence only offers binary outcomes for the pairwise probabilities. By increasing the value of $\gamma$, we begin to operate outside the saturation region, enabling richer information to be conveyed in the pairwise probabilities. Note, as $\gamma\rightarrow\infty$, we approach all probabilities equally a value of 0.5, which is also a loss of information. Hence, it is important to select a value of $\gamma$ that pushes the probabilities outside the saturation region but avoids all the probabilities concentrating at 0.5.

\section{Relationship between PoE-BT and Thurstone-Mosteller}
\label{app:rel}

\begin{figure}[htbp!]
    \centering
\includegraphics[width=0.48\textwidth]{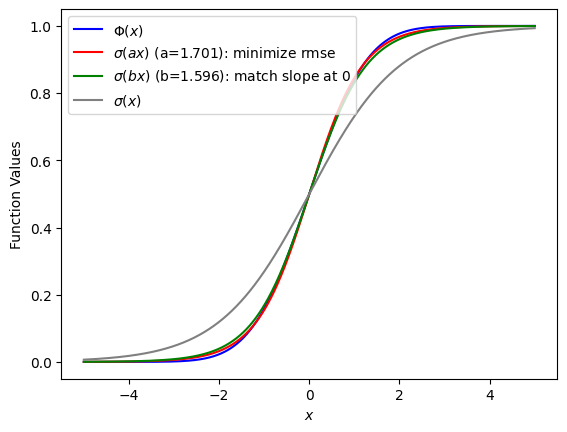}
    \caption{Linear mapping between $\sigma$ and $\Phi$.}
    \label{fig:linear_map}
\end{figure}

We argue that a linear scaling of the scores, $\gamma$, enables approximate mapping between the absolute scores output by various choice of $f$ (for example, it is empirically observed that scaling the argument of $\sigma(x)$ by 1.701 matches $\Phi(x)$ when minimizing rmse between the two functions. See Figure \ref{fig:linear_map} that shows a linear scaling between sigmoid and the cumulative normal distribution function. 

\begin{figure}[htbp!]
    \centering
\includegraphics[width=0.48\textwidth]{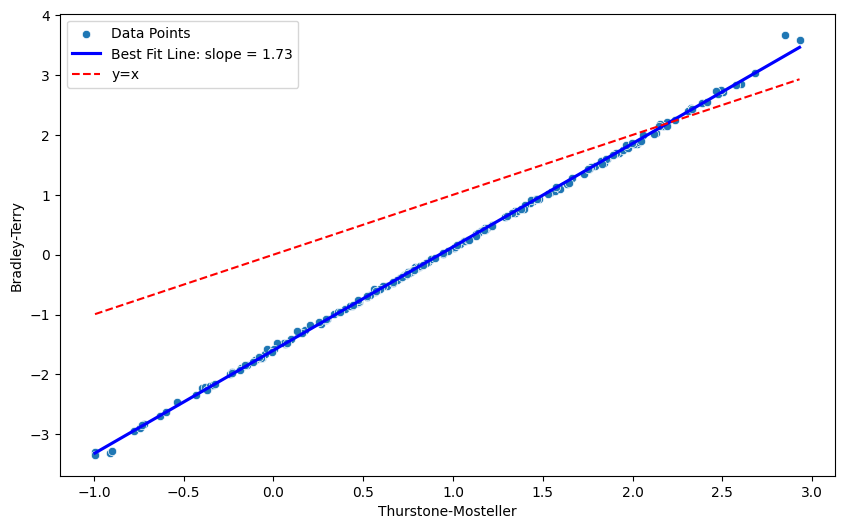}
    \caption{Relationship of scores (from zero-shot GPT-4o mini) using POE-BT and POE-TM for response time estimation.}
    \label{fig:proof}
\end{figure}

Hence, applying PoE-BT and Thurstone-Mosteller for comparative assessment can expect a linear scaling between their scores. Figure \ref{fig:proof} plots the score prediction from PoE-BT to the score prediction from Thurstone-Mosteller where the comparisons are generated by GPT4o mini for response time estimation. It is clear that 1.7 is a reasonable linear scaling between the scores from each of these methods.

\section{Prompts}
\label{app:prompts}

\begin{table}[htbp!]
\centering
\small 
\begin{tabular}{l|p{5cm}}
\toprule
Task & Prompt \\
\midrule
Response time & Which reading comprehension question can expect a longer candidate response time, 1 or 2? Return only 1 or 2. \\
\midrule
Difficulty & Which reading comprehension question is more difficult, 1 or 2? Return only 1 or 2. \\
   \bottomrule
    \end{tabular}
\caption{Prompts used for comparative training and assessment for each task type.}
\label{tab:prompts}
\end{table}

In this work, for comparative assessment, we consider two types of tasks: response time estimation and difficulty estimation. Table \ref{tab:prompts} summarizes the prompts for each task type.

\section{Hyperparameter tuning}
\label{app:hyp}

For the Llama-3.1-8B solution, for soft finetuning, we find $\gamma=5.0$ gives us the best results with hyperparameter finetuning for $\gamma\in\{0.1,0.5,1.0,2.0,5.0,10.0\}$. We apply parameter efficient finetuning using quantized low rank adaptation (QLoRA) \citep{dettmers2023qlora} for both the hard and soft finetuning involving 1 epoch with a batch size of 2, 50K pairwise examples, learning rate 1e-4 and QLoRA $\alpha=16$, $r=8$. Each model with 50K examples takes 13 hours to train on an NVIDIA A100 GPU.

For GPT4o mini hard finetuning, the training is performed for 1 epoch, 50K paired examples, learning rate multiplier
of 1.8 and batch size 33.

\section{USMLE difficulty estimation}
\label{app:usmlediff}

The USMLE dataset has been additionally annotated with difficulty scores. These annotations appear to be noisier, so we do not include them in main paper results. However, similar trends are observed from Table \ref{table:results_usmlediff} as was observed on the main paper comparative assessment tasks. Table \ref{tab:usmle_diff_baselines} further demonstrates that we achieve state-of-the-art performance for difficulty estimation on this task when comparing against the solutions submitted to the BEA shared task 2024 \citep{tack2024itec, felice2024british, duenas2024upn}.

\begin{table}[htbp!]
\centering
\small 
\begin{tabular}{cc|ccc}
\toprule
Model & Mode  & 
$\rho$ $(\uparrow)$ & $r$ $(\uparrow)$ & rmse $(\downarrow)$ \\
\midrule
    \multirow{2}{*}{GPT4o mini} 
        &  zero-shot &  7.5 & 5.8 & 0.310  \\
      &  hard & 32.9 & 34.7 & 0.291 \\
      \bottomrule
    \end{tabular}
\caption{Results using PoE-BT for USMLE difficulty estimation task using a full-set of comparisons.}
\label{table:results_usmlediff}
\end{table}

\begin{table}[htbp!]
\centering
\small 
\begin{tabular}{l|c}
\toprule
Approach & rmse $(\downarrow)$ \\
\midrule
Dummy Regressor Baseline  & 0.311 \\
EduTec: Electra & 0.299 \\
UPN-ICC & 0.303 \\
EduTec: Roberta & 0.304 \\
ITEC: Random Forest & 0.305 \\
Ours: comparative  & \textbf{0.291} \\
   \bottomrule
    \end{tabular}
\caption{Our best implementation against existing baselines for USMLE difficulty estimation.}
\label{tab:usmle_diff_baselines}
\end{table}

\section{Additional details}

We additionally trained an absolute (not pairwise) model with a regression loss function for USMLE response time estimation. This system achieved an rmse score of 26.1, which was not competitive with our equivalent comparative system.

Second, from \cite{liusie2024efficient}, product of experts with Gaussian experts is considered as a comparative scoring method. Theoretically, it is possible to finetune comparative LLM systems under the PoE framework applied to Gaussian experts. This would entail deducing training probabilities for the set of items in a training batch collectively. However, practically this was not feasible as our compute resources limited our training to a batch size of 2.

\section{Licenses}

For CMCQRD, the license \footnote{Available at \url{https://englishlanguageitutoring.com/datasets/cambridge-multiple-choice-questions-reading-dataset}} states the licensed dataset can be used for non-commercial research and educational purposes only. The USMLE dataset is distributed through the BEA shared task 2024.

\end{document}